\documentclass{article}

\PassOptionsToPackage{numbers, compress}{natbib}

\usepackage[preprint]{neurips_2024}

\usepackage[utf8]{inputenc} 
\usepackage[T1]{fontenc}    
\usepackage{hyperref}       
\usepackage{url}            
\usepackage{booktabs}  
\usepackage{multirow} 
\usepackage{amsfonts}       
\usepackage{nicefrac}       
\usepackage{microtype}      
\usepackage{xcolor}         
\usepackage{graphicx}
\usepackage{wrapfig}

\title{Dimba: Transformer-Mamba Diffusion Models}
\author{%
Zhengcong Fei, Mingyuan Fan, Changqian Yu, Debang Li
\\
\textbf{Youqiang Zhang, Junshi Huang}\thanks{Corresponding author} \\
Kunlun Inc.\\
Beijing, China\\
{\tt\small \{feizhengcong\}@gmail.com}
}

\begin{document}

\maketitle

\begin{abstract}

This paper unveils Dimba, a new text-to-image diffusion model that employs a distinctive hybrid architecture combining Transformer and Mamba elements. Specifically, Dimba sequentially stacked blocks alternate between Transformer and Mamba layers, and integrate conditional information through the cross-attention layer, thus capitalizing on the advantages of both architectural paradigms. We investigate several optimization strategies, including quality tuning, resolution adaption, and identify critical configurations necessary for large-scale image generation. The model's flexible design supports scenarios that cater to specific resource constraints and objectives. 
When scaled appropriately, Dimba offers substantial throughput and a reduced memory footprint relative to conventional pure Transformers-based benchmarks. Extensive experiments indicate that Dimba achieves comparable performance compared with benchmarks in terms of image quality, artistic rendering, and semantic control. We also report several intriguing properties of architecture discovered during evaluation and release checkpoints in experiments. 
Our findings emphasize the promise of large-scale hybrid Transformer-Mamba architectures in the foundational stage of diffusion models, suggesting a bright future for text-to-image generation.
Project page: \url{https://dimba-project.github.io/}.
\end{abstract}

\section{Introduction}
Diffusion models create data from noise \cite{song2020score}, and are trained on the inversion of data pathways that lead towards random noise. By leveraging the approximation and generalization properties inherent in neural networks, these models can generate novel data points not found within the training set but still aligned with its distribution \cite{sohl2015deep,song2019generative,yang2023diffusion}.
This generative modeling technique has proven highly effective for high-dimensional perceptual data, such as images \cite{ho2020denoising,nichol2021improved,song2020denoising}.
In recent years, text-to-image diffusion models, such as DALL-E\cite{ramesh2022hierarchical,betker2023improving}, Imagen \cite{saharia2022photorealistic}, Stable Diffusion \cite{rombach2022high,podell2023sdxl,esser2024scaling}, EMU \cite{dai2023emu}, and so on \cite{balaji2022ediff,chen2023pixart,chen2024pixart,li2024playground,zheng2024cogview3}, have revolutionized the field of photorealistic image synthesis, dramatically impacting applications including image editing \cite{couairon2022diffedit,kim2022diffusionclip,brooks2023instructpix2pix,kawar2023imagic,yang2023paint}, video generation \cite{wu2023tune,ho2022video,ho2022imagen,mei2023vidm,blattmann2023stable}, 3D assets creation \cite{poole2209dreamfusion,lin2023consistent123}, etc. 
Prediction of noise or previous states in the reverse
diffusion process usually uses U-Net architectures based on a convolutional neural network (CNN)  \cite{ho2020denoising,song2019generative} or Transformer \cite{peebles2023scalable,yang2022your,bao2023all}, as highlighted in prior studies.

\begin{figure}[!ht]
\centering
\includegraphics[width=1\linewidth]{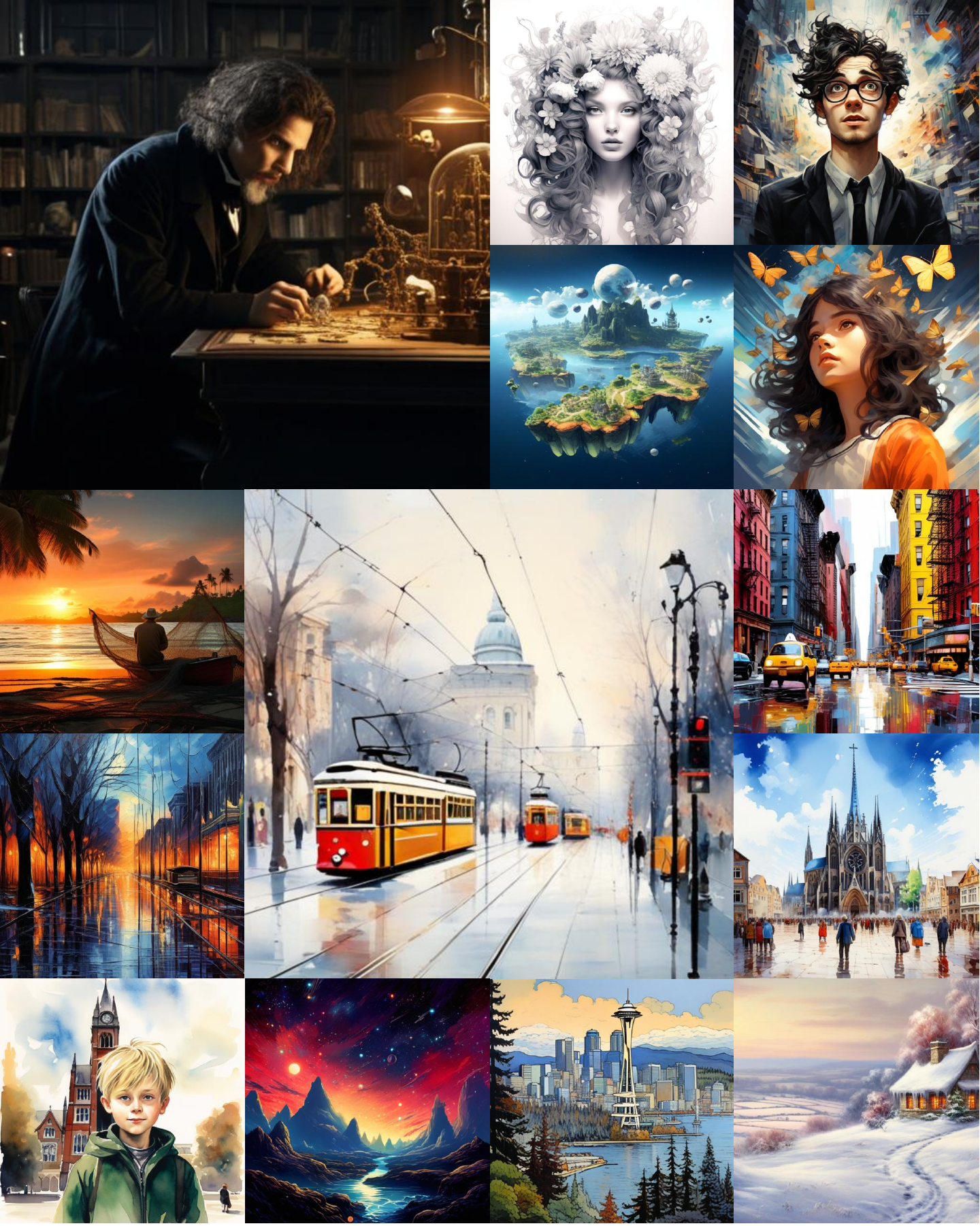}
\caption{\textbf{Images generated from Dimba. } The Dimba model can output high aesthetic, objective natural and consistent, and follow user textual instructions.}
\label{fig:teaser}
\end{figure}

On the other hand, advancements in state space models (SSM) have paved the way for a new era in achieving a balance between computational efficiency and model flexibility. Originating from the
classic Kalman filter model \cite{kalman1960new}, various SSM-based approaches \cite{gu2021combining,gu2021efficiently,gupta2022diagonal,gu2020hippo}, have been proposed to process sequence data, proving highly effective in managing long-distance dependencies across diverse tasks and modalities. Their efficiency in processing extended sequences is attributed to convolutional computations and nearly linear computational complexity. More recently, Mamba \cite{gu2023mamba} advances by incorporating time-varying parameters into the SSM and proposing a hardware-aware algorithm that facilitates highly efficient training and inference. 
The superior scaling performance of Mamba positions it as a promising alternative to Transformer \cite{vaswani2017attention} in language modeling. Additionally, a series of SSM-based backbone networks \cite{zhu2024vision,yang2024vivim,patro2024simba} have been explored for processing visual data, such as images and videos \cite{yan2023diffusion,fei2024scalable,hu2024zigma,fei2024diffusion}.
It becomes imperative to conduct an in-depth analysis of how Transformer-Mamba mixed structures are applied to diffusion-based multimodal generation tasks and identify the essential techniques to ensure model capacity.

In this paper, we introduce Dimba, a new diffusion model design for text-to-image synthesis. Specifically, Dimba is based on a hybrid architecture, which combines Transformer layers with Mamba layers, a recent advanced state-space model. 
Despite the immense popularity of the CNN or Transformer as the predominant architecture for diffusion models, it still exhibits substantial drawbacks. 
Particularly their extensive memory and computational requirements, which limit their wide application in handling long contexts without special designs. In contrast,  recurrent-based neural network models, which summarize an arbitrarily long context in a single hidden state, can reduce these issues. 
Dimba thus combines two orthogonal architectural designs, enhancing performance and throughput while maintaining a manageable memory footprint. Note that the flexibility of the Dimba architecture allows it to accommodate various design choices based on hardware and performance needs. Several generated results are illustrated in Figure \ref{fig:teaser}.
Extensive experiments indicate that Dimba, across different parameter settings, provides superior image quality and semantic alignment compared to existing mainstream diffusion models, and its performance on T2I-CompBench \cite{huang2023t2i} also underscores its strength in semantic control. 
The core contributions of our work are: 
\begin{itemize}
    \item We present Dimba, a novel diffusion architecture that combines attention and Mamba layers, for text-to-image generation. We showed how Dimba provides flexibility for balancing throughput and memory requirements, while maintaining a comparable performance.
\item We curate a large-scale high-quality image-text dataset emphasizing aesthetically superior images. These images are evaluated using a score network and paired with dense, precise captions generated by advanced auto-label models. It benefits from sufficient text-image semantic alignment as well as image aesthetic quality improvement. 
\item We train Dimba in a staged progressive strategy, that is first pre-training model with large-scale data and then adaption to high-resolution as well as quality tuning with a small well-selected set. 
Our efforts aim to provide valuable insights into efficient diffusion models, assisting individual researchers and startups in developing high-quality text-to-image models at reduced memory costs.
\end{itemize}

\section{Related Works}

\paragraph{Image generation with diffusion models.} 

There have been substantial advancements in the field of image generation in recent years, particularly through diffusion process \cite{rombach2022high,saharia2022photorealistic,fei2022progressive,ramesh2022hierarchical,zhang2023adding,fei2024music,fei2023gradient}. 
This technique typically involves starting with Gaussian noise and iteratively refining it through a series of steps until it aligns with the target distribution. It has shown remarkable capabilities, usually surpassing the performance of GAN \cite{goodfellow2020generative,fei2023masked} and VAE \cite{kingma2013auto}-based models. 
To generate image conditioned on text, several approaches have been introduced, such as GLIDE \cite{nichol2021glide}, DALLE \cite{ramesh2022hierarchical}, Imagen \cite{ho2022imagen}, LDM \cite{rombach2022high} and so on \cite{balaji2022ediff,chen2023pixart,chen2024pixart,li2024playground,zheng2024cogview3}. 
These models train diffusion models using large-scale text-image pairs, enabling the creation of images based on text conditions. 
In between, U-Net is responsible for predicting noise based on the textual embedding in latent space. 
The CNN-based UNet is characterized by a group of down-sampling blocks, a group of up-sampling blocks, and long skip connections between the two groups \cite{dhariwal2021diffusion,ramesh2022hierarchical,rombach2022high,saharia2022photorealistic}. In contrast, Transformer-based architecture replaces part of sampling blocks with self-attention while keeping the remain untouched \cite{peebles2023scalable,yang2022your,bao2023all}. SSM-based UNet \cite{yan2023diffusion,fei2024scalable,hu2024zigma,zhu2024dig} within image sequences have also shown promising results with computation optimization. 
In this paper, we try to combine Transformer with Mamba \cite{gu2023mamba} to trait a better computation efficiency as well as comparable performance.

\paragraph{State space models for visual applications.}
SSM is employed for sequential modeling, with one-dimensional inputs and outputs, and has been applied in control theory, signal processing, and natural language process \cite{hamilton1994state}.
A series of studies have focused on structural improvements \cite{gu2021efficiently,smith2024convolutional,fu2022hungry,mehta2022long}.
Recently, Mamba \cite{gu2023mamba} was proposed with a data-dependent SSM layer and built a generic language model backbone, which outperforms Transformers at various sizes on large-scale real data and enjoys linear scaling in sequence length \cite{nguyen2022s4nd,islam2023efficient,wang2023selective}.  
Regarding the integration of attention mechanisms with SSMs, \cite{zuo2022efficient} combines an S4 layer \cite{gu2021efficiently} with a local attention layer, and demonstrates effectiveness on small models and simple tasks.  \cite{pilault2024block} starts with an SSM layer followed by chunk-based Transformers, showing improved perplexity metrics. \cite{fathullah2023multi} adds an SSM layer before the self-attention in a Transformer layer, while \cite{saon2023diagonal} adds the SSM after the self-attention, both yielding improvements in speech recognition. \cite{park2024can} replaces the MLP layers in the Transformer by Mamba layers, and shows benefits in specific tasks. \cite{yan2023diffusion,fei2024scalable,ma2024u} supplants attention mechanisms with a more scalable SSM-based backbone to generate high-resolution images. 
Most similar work \cite{lieber2024jamba} scaling the mixed mamba and Transformer structure in causal language modeling while our works focus on bi-direction structure for text-to-image synthesis.

\section{Methodology}

\subsection{Model Architecture}
 \begin{wrapfigure}{r}{0.55\textwidth}
  \centering
  \includegraphics[width=0.55\textwidth]{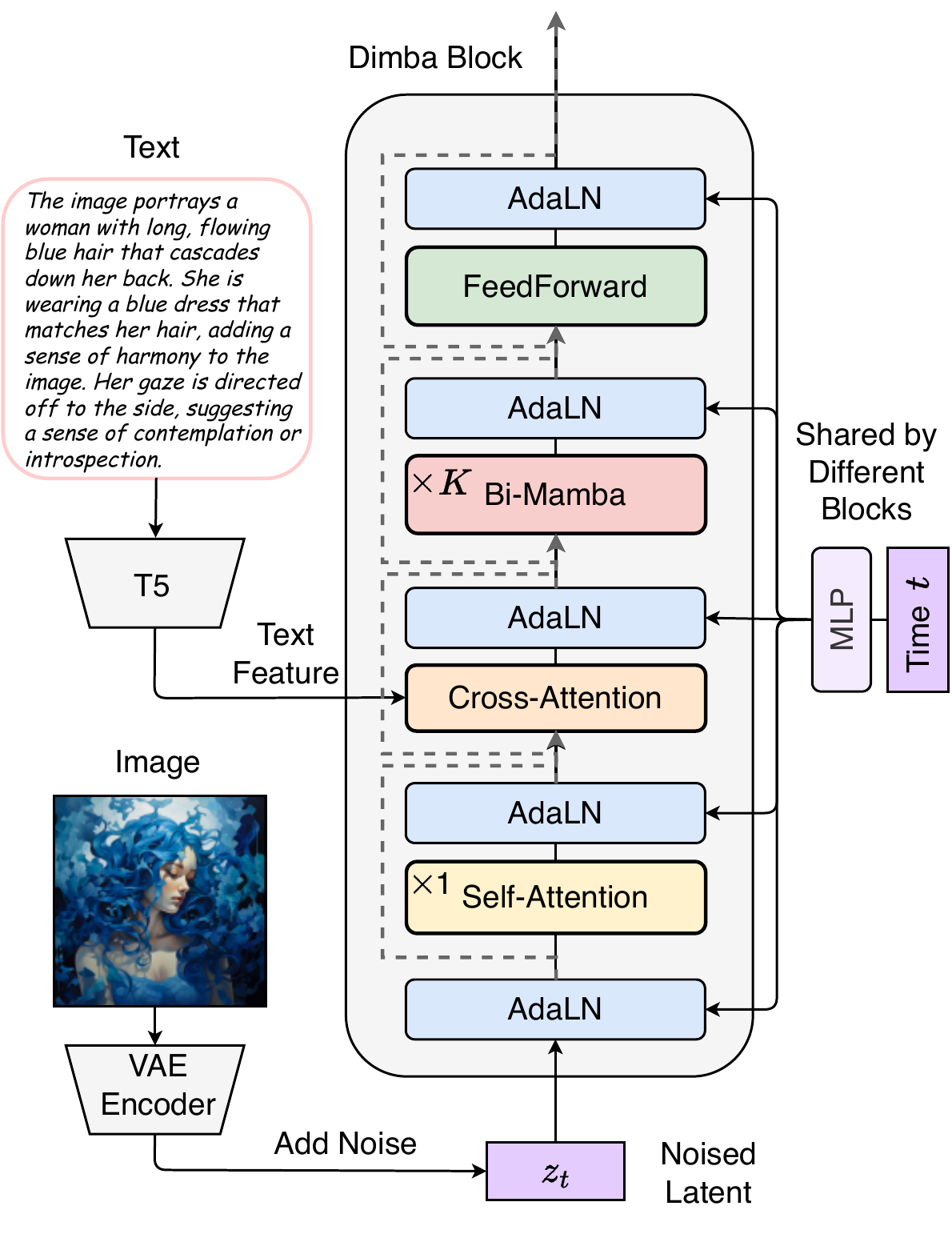}
  \caption{\textbf{Model architecture of Dimba}. Mamba and Transformer layers are interleaved in a stacked manner. The text feature is incorporated with a cross-attention layer. Time information is projected with shared MLP before inserting to different AdaLN layers.} 
  \label{fig:pipeline-model}
\end{wrapfigure}

Generally, Dimba is a hybrid diffusion architecture that mixes a series of Transformer layers with Mamba layers, complemented by a cross-attention module for managing conditional information.  
We call the combination of these three elements as a Dimba block. See Figure \ref{fig:pipeline-model} for an illustration. 
Prior experiments indicate that directly concatenating textual information with prefix embeddings can decelerate training convergence. 
The integration of Transformer and Mamba elements in Dimba offers flexibility in balancing the sometimes conflicting goals of low memory usage, high throughput, and high quality. A significant consideration is the scalability of Transformer models to long contexts, where the quadratic growth of the memory cache becomes a limiting factor. By substituting some attention layers with Mamba layers, the overall cache size is reduced. Dimba's architecture is designed to require a smaller memory cache compared to a conventional Transformer.
In terms of throughput, attention operations consume a relatively small portion of the inference and training FLOPS for short sequences. However, when sequences keep increasing, attention operations dominate the computational resources. Conversely, Mamba layers are more compute-efficient, and increasing their ratio enhances throughput, particularly for longer sequences.

For implementation specifics, the fundamental unit, Dimba block, which is repeated in sequence, the hidden states output from the previous layer serves as the input for the next layer. Each such layer contains an attention and a Mamba module, \emph{i.e.}, we set the attention-Mamba ratio to 1:$K$, followed by a multi-layer perceptron (MLP). Drawing from  \cite{chen2023pixart}, we incorporate a global shared MLP and layer-wise embedding for time-step information in AdaLN. 
Note that Mamba layers in each Dimba block also incorporate several adaptive normalization layers (AdaLN) as \cite{peebles2023scalable} that help stabilize training in large-scale models.

\subsection{Dataset Construction}\label{sec:data}
\begin{figure}[t]
\centering
\includegraphics[width=1\linewidth]{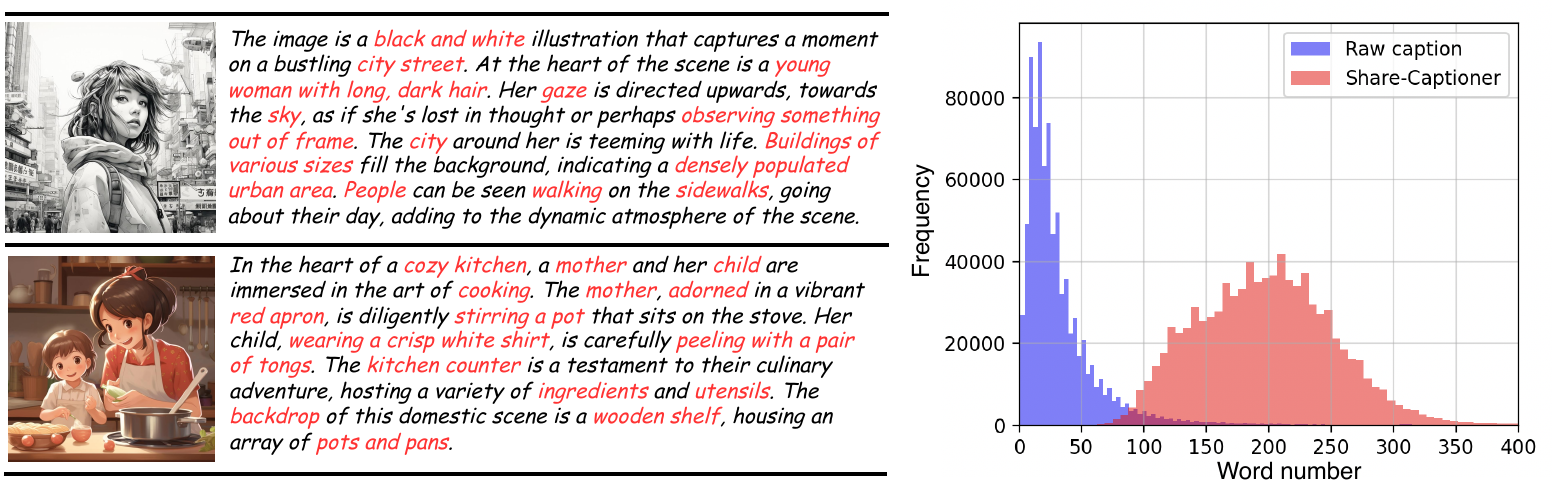}
\caption{\textbf{Training data illustration and histogram visualization of the caption length. } (a) Auto-labeling caption provides accurate textual descriptions for images, and we outline the valid nouns and verbs in red color; (b) We randomly select 1M captions from the raw captions and re-labeled captions to draw the corresponding histogram. }
\label{fig:text-image-samples}
\end{figure}

\paragraph{Image-text pair auto-labeling and quality estimation.}
The captions in mainstream image-text datasets sourced from web crawls often suffer from issues such as text-image misalignment, incomplete descriptions, hallucinations, and infrequent vocabulary, as demonstrated \cite{chen2023pixart,fei2022deecap,chen2023sharegpt4v,yu2023capsfusion}. 
Moreover, it is worth noting that the public dataset, such as LAION-400M \cite{schuhmann2021laion}, predominantly contains simplistic product previews from websites, which are suboptimal for training text-to-image generation models aimed at seeking diversity in object combinations.
Additionally, previous studies indicate that images with higher artistic styles are generally preferred by users \cite{li2024playground}.
To address these problems and generate captions with high accuracy and information density, we construct our training dataset by first crawling a large-scale internal dataset and then filtering high-quality images according to automatic scorer, to enhance the aesthetic quality of generated images beyond realistic photographs.
In between, we use LAION-Aesthetics-Predictor V2 from \footnote{https://laion.ai/blog/laion-aesthetics/} for image quality estimation, which integrates CLIP and MLP models trained on datasets with specific aesthetic ratings. 
Finally, we employ the state-of-the-art image captioning model, i.e., ShareCaptioner ~\cite{chen2023sharegpt4v}, available from open-source repository\footnote{https://huggingface.co/spaces/Lin-Chen/Share-Captioner}, to relabel the data. Using the prompt, ``\textit{Analyze the image in a comprehensive and detailed manner.}'', we significantly improve the quality of captions compared to the original ones, as shown in Figure~\ref{fig:text-image-samples} (a).

\paragraph{Dataset composition analysis.}
It is believed that effective training strategies can yield robust text-to-image models even with limited data \cite{li2024playground,chen2023pixart}. 
Following this principle, we have compiled an internal dataset of approximately 43 million images. Besides, we further filter out balanced high-score images with shortest edge $>1024$ resolution, yielding a high-refined quality tuning dataset of 600K images. 
Following \cite{chen2024pixart}, we also conduct a comprehensive analysis of the collected dataset, including the caption length and vocabulary analysis with NLTK package \cite{loper2002nltk}. Here we define the valid distinct nouns as those appearing more than 10 times in the dataset. 
As shown in Figure \ref{fig:text-image-samples}(b), the average caption length increased significantly to 185 words, highly enhancing the descriptive power of the captions. Additionally, we use the T5 encoder with 350 tokens maximum length for sufficient token processing.  
Lastly, the dataset also ensures sufficient valid nouns and average information density for training. Specifically, Share-Captioner labeled captions significantly increase the valid ratio, from 15.5\% to 27.2\%, and average noun count per image, from 10.5 per image to 21.2 per image, improving concept density. 
To enhance textual diversity and mitigate biases from generative captions, the Dimba model was trained using a combination of long (Share-Captioner) and short (Raw) captions in 90\% to 10\%.

\subsection{Training Strategy}

\paragraph{Quality tuning.} 

High-aesthetic images, such as the 600K images we collected with automatic quality evaluation, often share common statistical attributes as noted by  \cite{ruiz2023dreambooth,wallace2023diffusion}. 
We hypothesize that a well-trained text-to-image diffusion model has the potential to create highly aesthetic images, but the generation process is not consistently guided toward producing images with these desired statistics. Therefore, we employ quality tuning \cite{dai2023emu}, a technique that effectively restricts outputs to a high-quality subset. 
Specifically, we first pre-train Dimba model with the complete image-text data and subsequently fine-tune it with a small batch size and high-quality dataset set. 
It is crucial to apply early stopping, as extended training on a small dataset can lead to overfitting and a loss of generalization in visual concepts. 
In practice, we fine-tune for every 10K iteration and select the best ckpts in human observation empirically, despite the continued loss decreasing.

\paragraph{Adapting to higher resolution.} 
To assess the advantages of hybrid architecture, we investigate expanding the generated image resolution of Dimba model.
When transitioning from short input sequence to longer processing, previous works \cite{lieber2024jamba} suggest that explicit positional information may not be required for the hybrid mamba-attention architecture. Presumably, the Mamba layers, which are placed before attention layers, provide implicit position embedding (PE) information.
Herein, we examine the feasibility of employing the PE Interpolation trick \cite{xie2023difffit,chen2024pixart}, which involves initializing the PE of a high-resolution model by interpolating those of a low-resolution model, thereby potentially enhancing the initial state of the former and expediting subsequent fine-tuning processes. 
Our findings indicate that leveraging the PE interpolation technique leads to notable improvements in model convergence and image quality. The fine-tuning quickly converges at 20K steps, and further training marginally improves the performance. When dropping out the PE interpolation trick, the Dimba model generally results in similar performance after 50K steps. 
Consequently, combining the PE Interpolation trick with Mamba implicit position modeling enables rapid convergence of higher resolution generation, and alleviates the necessity of training models from scratch to achieve high-resolution outputs.

\begin{table}[t]
\centering
\caption{\textbf{Image quality comparison between Dimba with recent text-to-image models}. We consider several essential factors including model size, the total volume of training images, COCO FID-30K scores in zero-shot, and the computational cost in GPU days. 
The proposed highly effective approach significantly reduces resource consumption, including training data usage and training time. 
The baseline data is sourced from \cite{chen2023pixart}. “+” in the table denotes an unknown internal dataset size.}
\label{tab:modelsize_params_mscocofid}
\resizebox{0.8\linewidth}{!}{ 
\begin{tabular}{lccccc}
        \toprule
    	 Method &  Type &  ${\#}$Params &  ${\#}$Images  &  FID-30K$\downarrow$ &  GPU days \\
    	\midrule
    	DALL·E & Diffusion   & 12.0B & {250M} & 27.50 & - \\
            GLIDE  & Diffusion   & 5.0B  & {250M} & 12.24 & - \\
            LDM    & Diffusion   & 1.4B  & {400M} & 12.64 & -\\
            DALL·E 2 & Diffusion & 6.5B  & {650M} & 10.39 & 41,667 A100\\
            SDv1.5 & Diffusion   & 0.9B  & {2000M} & 9.62  & 6,250 A100\\
            GigaGAN & GAN   & 0.9B  & {2700M} & 9.09  & 4,783 A100 \\
            Imagen & Diffusion   & 3.0B  & {860M} & 7.27  & 7,132 A100\\
            RAPHAEL & Diffusion  & 3.0B  & {5000M+} & 6.61  & 60,000 A100 \\
    	PixArt & Diffusion   & {0.6B} & {{25M}} & 7.32 & {753 A100} \\ \midrule 
     Dimba-L & Diffusion & 0.9B &43M & 8.93 & 704 A100 \\
     Dimba-G & Diffusion  & 1.8B &43M & 8.15 & 1,980 A100 \\
        \bottomrule

\end{tabular}
}
\end{table}

\section{Experiments}

This section begins by delineating the intricacies of the training and evaluation protocols. Following this, we offer exhaustive analyses across primary metrics from image quality and image-text alignment. Subsequently, we delve into the pivotal design elements incorporated within Dimba to enhance both its efficiency and effectiveness, as evidenced by ablation studies. Finally, we showcase some generated results with popular benchmarks intuitively.

\subsection{Experimental Settings}

\paragraph{Implementation details.} 
We follow prior works \cite{saharia2022photorealistic,podell2023sdxl,chen2024pixart,betker2023improving} by utilizing the T5 large language model, \emph{i.e.}, 4.3B Flan-T5-XXL\footnote{https://huggingface.co/google/flan-t5-xxl}, as the text encoder for conditional text feature extraction, and used standard transformer-mamba \cite{lieber2024jamba} mixed models as our base network architecture. 
We adapt the length of extracted text tokens to 350, aligning it with the distribution of auto-labeled caption lengths. This adjustment accommodates the denser captions curated within Dimba, facilitating the provision of finer-grained details. For the XL version, we incorporate 28 Dima blocks with a hidden size of 1152, while for the Giant size version, we utilize 40 Dima blocks with a hidden size of 1408. The attention-to-Mamba ratio is set to $K=1$, and the patch size is standardized to 2. 
To capture the latent features of input images, we employ a pre-trained and frozen VAE from SDXL \citep{rombach2022high,podell2023sdxl}. Before feeding the images into the VAE, we resize and center-crop them to have the same size. We leave the multi-aspect augmentation to enable aspect image generation in future works. 
we use CAME optimizer \cite{luo2023came} with a weight decay of 0 and a constant learning rate of 2e-5, instead of the regular AdamW \cite{loshchilov2017decoupled} optimizer, motivated by the advantage to reduce the dimension of the optimizer's state, thereby minimizing GPU memory usage without compromising performance.
For the sampling algorithm, to optimize computational efficiency, we chose to employ the DPM-Solver \cite{lu2022dpm} with 20 inference steps by default.

\paragraph{Evaluation metrics.}  We comprehensively evaluate the generative performance of Dimba via three primary metrics, i.e., Fréchet Inception Distance~(FID)~\citep{heusel2017gans} on MSCOCO dataset~\citep{lin2014microsoft}, compositionality on T2I-CompBench~\citep{huang2023t2i}, and human-preference rate on user study.

\subsection{Performance Comparisons} 

\paragraph{Image quality assessment.} 
The FID score serves as a widely used metric for assessing the quality of generated images relative to ground-truth images from disparate distributions. A comparative analysis between our Dimba and other prominent text-to-image methods, with respect to FID scores and their corresponding training durations is summarized in Table~\ref{tab:modelsize_params_mscocofid}. 
Generally, when tested for zero-shot performance on the COCO dataset, Dima of the large version achieves an FID score of 8.93. It is particularly notable as it is accomplished in merely 11.2\% of the training time, i.e., {704} vs 6250 A100 GPU days, and merely {2.0\%} of the training samples, i.e., 43M vs {2B} images, relative to the benchmarks.
Moreover, it achieves comparable performance with less training time, compared with PixArt which is based on a pure Transformer for efficient training. 
Meantime, mixed structure provides a more flexible selection in light of inference resource requirements.
Encouragingly, compared to state-of-the-art methods typically trained using substantial resources, remarkably consumes approximately 2\% of the training resources while also achieving a similar FID performance. 
We posit that evaluations conducted by human users offer a more comprehensive assessment of image quality, an analysis that we undertake in subsequent sections. Additionally, as anticipated, scaling the model parameters consistently enhances image quality.

\paragraph{Image-text alignment assessment.} 
Similar to \cite{chen2023pixart}, we also scrutinize the congruence between generated images and textual conditions utilizing T2I-Compbench~\citep{huang2023t2i}, which serves as a comprehensive benchmark designed to evaluate the compositional text-to-image generation capability. 
As delineated in Table~\ref{benchmark:t2icompbench}, we assess several critical dimensions, including attribute binding, object relationships, and intricate compositions. The findings reveal that Dimba-G demonstrates exemplary performance across a myriad of evaluation metrics. This noteworthy achievement can predominantly be attributed to the meticulous curation of the dataset, which facilitated the utilization of high-quality text-image pairs, thus enhancing alignment capabilities.
It is worth noting that as the quality-tuning process incorporates more aesthetically-oriented data, Dimba tends to generate images with a more cohesive stylistic resemblance.

\begin{figure}[!ht]
\centering
\includegraphics[width=0.98\linewidth]{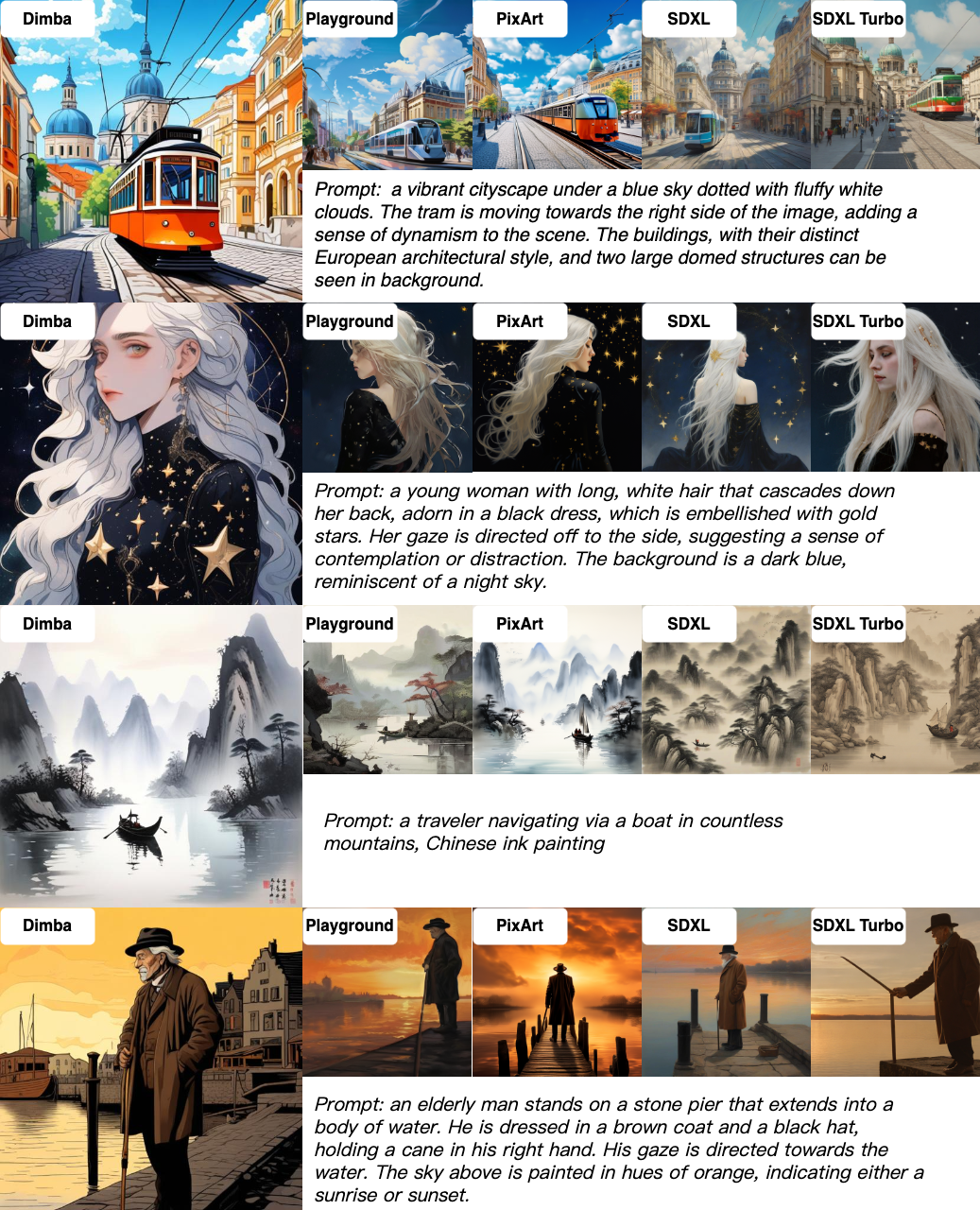}
\caption{\textbf{Qualitative comparison of Dimba with four other open-source text-to-image models}. Baselines include Playground v2.5, PixArt, SDXL and SDXL Turbo. Images generated by Dimba are very competitive with these benchmarks and show more details and aesthetics.}
\label{fig:case}
\end{figure}

\begin{table}[t]
\centering
\caption{\textbf{Image-text alignment results on T2I-CompBench.} We can see that Dimba-G demonstrated exceptional performance in attribute binding, object relationships, and complex compositions, indicating our method achieves comparable compositional generation ability with well-designed datasets and staged training. The baseline data are sourced from ~\cite{huang2023t2i}.} 
\label{benchmark:t2icompbench}
\resizebox{0.95\linewidth}{!}{ 
\begin{tabular}
{lcccccc}
\toprule
\multicolumn{1}{c}
{\multirow{2}{*}{ Model}} & \multicolumn{3}{c}{ Attribute Binding } & \multicolumn{2}{c}{ Object Relationship} & \multirow{2}{*}{ Complex$\uparrow$}
\\
\cmidrule(lr){2-4}\cmidrule(lr){5-6}

&
{ Color $\uparrow$ } &
{ Shape$\uparrow$} &
{ Texture$\uparrow$} &
{ Spatial$\uparrow$} &
{ Non-Spatial$\uparrow$} &
\\
\midrule
Stable v1.4  & 0.3765 & 0.3576 & 0.4156 & 0.1246 & 0.3079 & 0.3080  \\
Stable v2 & 0.5065 & 0.4221 & 0.4922 & 0.1342 & 0.3096 & 0.3386  \\
Composable v2 & 0.4063 & 0.3299 & 0.3645 & 0.0800 & 0.2980 & 0.2898  \\
Structured v2 & 0.4990 & 0.4218 & 0.4900 & 0.1386 & 0.3111 & 0.3355  \\
Attn-Exct v2 & 0.6400 & 0.4517 & 0.5963 & 0.1455 & 0.3109 & 0.3401  \\
GROS &0.6603 & 0.4785& 0.6287 &0.1815 &0.3193 & 0.3328\\
Dalle-2 &0.5750 &0.5464 & 0.6374 &0.1283 & 0.3043 &0.3696 \\
SDXL & 0.6369& 0.5408 &0.5637 &0.2032 &0.3110 &0.4091\\
PixArt &0.6886 &0.5582 &0.7044&0.2082 &0.3179 &0.4117 \\ \midrule 
Dimba-G &0.6921&0.5707&0.6821 &0.2105 & 0.3298 & 0.4312 \\
\bottomrule
\end{tabular}
}

\end{table}

\begin{figure}[t]
\begin{center}
\includegraphics[width=1\linewidth]{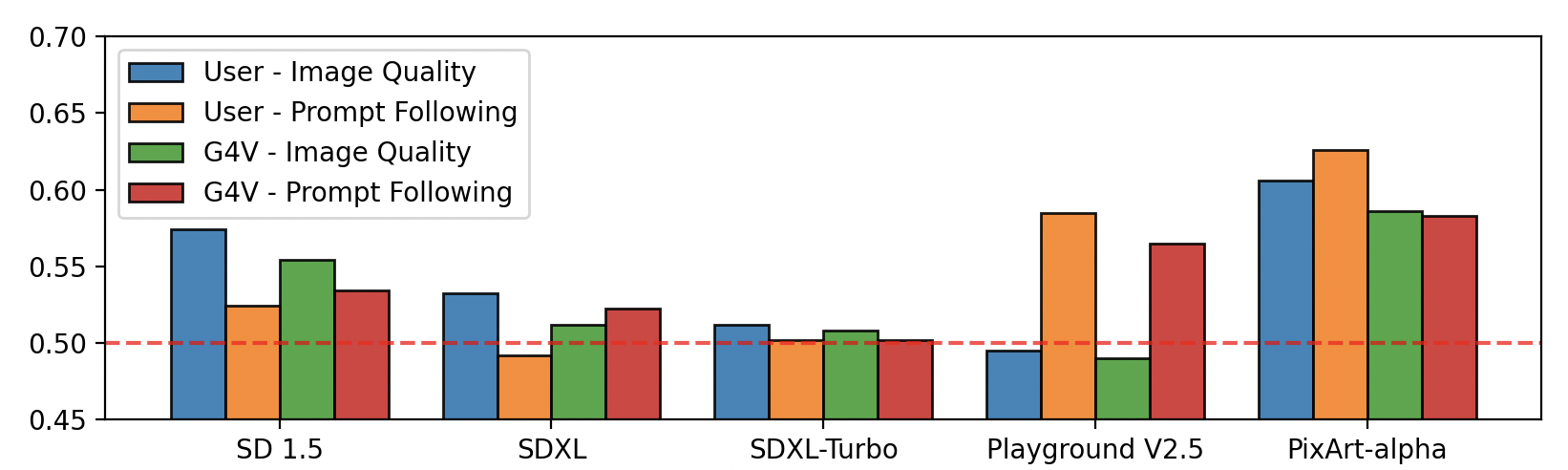}
\caption{\textbf{User study and AI preference on fixed prompts.} The ratio values indicate the percentages of participants preferring Dimba over the corresponding baselines. Dimba achieves a superior capacity in both image quality and prompt following.} 
\label{fig:user_study}
\end{center} 
\end{figure}

\paragraph{User study.} 

While quantitative evaluation metrics provide insights into the overall distribution of generated and ground-truth image sets, they may not offer an independent assessment of the visual quality of the images. To address this limitation, we conducted a user study to augment our evaluation and afford a more intuitive appraisal of Dimba's performance.
For this study, we selected the top-performing models, specifically SD 1.5, SDXL, SDXL-Turbo, Playground v2.5, and PixArt. Subsequently, we meticulously curated 200 prompts from publicly available papers and obtained detailed captions using GPT-4 with in-context settings. The human evaluation results are presented in Figure \ref{fig:user_study}. We observe that the majority of scores surpass the 0.5 threshold, indicated by the red line, signifying that Dimba-G achieves comparable synthesis performance with mainstream baselines.

\paragraph{AI preference.}

Additionally, we employ the advanced multimodal model, GPT-4 Vision, as the evaluator for AI preference study.  In each trial, we present GPT-4 Vision with two images: one generated by Dimba and another by a competing benchmark. We meticulously craft distinct prompts to guide GPT-4 Vision in voting based on both image quality and alignment with accompanying text. The outcomes of this study, as illustrated in Figure \ref{fig:user_study}, exhibit consistent findings with those of the human preference studies. 

\begin{figure}[t]
\begin{center}
\includegraphics[width=0.98\linewidth]{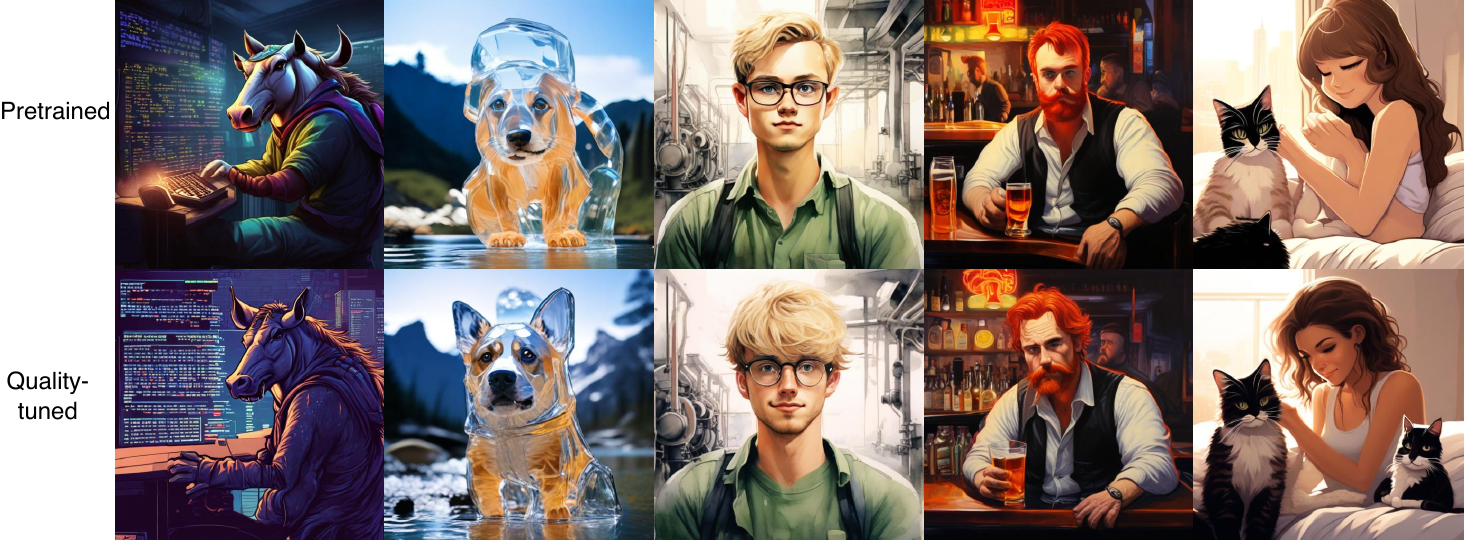}
\caption{\textbf{Comparison of images generated by the pre-trained and quality-tuned Dimba model.} Quality-tunning can significantly optimize the synthesis image in details and aesthetics.
}
\label{fig:quality-tune}
\end{center}
\end{figure}

\paragraph{Case study.}
First, we undertake a comparative analysis of our quality-tuned model, Dima, with the pre-trained model. See Figure \ref{fig:quality-tune} for random, not cherry-picked, qualitative examples before and after quality-tuning. It is important to note the inclusion of highly aesthetic non-photorealistic images, which validates our hypothesis that adhering to specific photography principles during the curation of the quality-tuning dataset enhances aesthetics across a diverse range of styles. 
Besides, we show more intuitive examples of generated images using Dima and baselines in Figure \ref{fig:teaser} and Figure \ref{fig:case}.

\section{Limitations}

Dimba model exhibits limitations attributable to biases present in the training data. Particularly, it struggles to generate specific styles, scenes, and objects, with notable challenges observed in text and hand generation. Imperfections persist in various aspects: complex prompts input by users may not be fully aligned, face generation may entail flaws, and the risk of generating sensitive content remains.
Furthermore, it is imperative to acknowledge the potential negative social ramifications associated with text-to-image models. Such models may inadvertently perpetuate stereotypes or discriminate against certain groups. Addressing these issues necessitates concerted efforts in future research endeavors.

\section{Conclusion}

This paper presents Dimba, a pioneering diffusion architecture tailored for text-to-image generation, which integrates attention and Mamba layers. We also present an implementation of Dimba, achieving performance on par with meticulously curated datasets. Furthermore, we elucidate how Dimba offers flexibility in balancing performance and memory requirements while maintaining high throughput. Through extensive experimentation, we explore various design considerations, including data construction, quality tuning, and resolution adaptation, shedding light on insights garnered during the developmental phase. These findings are poised to guide future endeavors in hybrid attention–Mamba backbone generation. To facilitate such research, we intend to release code and model checkpoints derived from smaller-scale training iterations.

\bibliographystyle{plainnat}
\bibliography{main}

\end{document}